%% file: acl_latex.tex
\pdfoutput=1

\documentclass[11pt]{article}

\usepackage{acl}
\usepackage{times}
\usepackage{latexsym}
\usepackage{multirow}

\usepackage[T1]{fontenc}

\usepackage[utf8]{inputenc}

\usepackage{microtype}
\usepackage{pifont}
\usepackage{booktabs}
\usepackage{amsfonts}
\usepackage{amsmath}
\usepackage{xcolor}
\usepackage{csquotes}
\usepackage{enumitem}
\usepackage{graphicx}
\usepackage{xspace}
\usepackage{hyperref}

\newcommand{\question}{q}
\newcommand{\gtanswer}{a}
\newcommand{\panswer}{\hat{\gtanswer}}
\newcommand{\dq}{x}
\newcommand{\da}{y}
\newcommand{\pda}{\hat{\da}}
\newcommand{\fone}{\phi}
\newcommand{\numdis}{n}
\newcommand{\knwldg}{e}
\newcommand{\numknwldg}{m}
\newcommand{\samplesize}{N}
\newcommand{\wiki}{W}
\newcommand{\context}{C}

\newcommand{\acc}{{\textsc{Acc}}}
\newcommand{\comp}{{\textsc{Comp}}}
\newcommand{\flue}{{\textsc{Flue}}}
\newcommand{\hovscore}{\textsc{HO}}

\newcommand{\dpr}{\textsc{DPR}}
\newcommand{\dprk}[1]{DPR@#1}
\newcommand{\jpr}{\textsc{JPR}}
\newcommand{\jprk}[1]{JPR@#1}

\newcommand{\asqa}{{\sc ASQA}\xspace}
\newcommand{\ambig}{{\sc AmbigQA\xspace}}
\newcommand{\msmarco}{{\sc MS-NLG}}
\newcommand{\eli}{{\sc ELI5}\xspace}
\newcommand{\nq}{{\sc NQ}}
\newcommand{\nqopen}{{\sc NQ-open}}
\newcommand{\squad}{{\sc SQuADv2}}
\newcommand{\narrative}{{\sc NarrativeQA}}
\newcommand{\coqa}{{\sc CoQA}}

\newcommand{\lrouge}{ROUGE-L\xspace}
\newcommand{\qaf}{{\footnotesize Disambig}-F1\xspace}
\newcommand{\strem}{STR-EM\xspace}
\newcommand{\ovscore}{DR\xspace}

\newcommand{\hpwc}{{\sc HP-w/-C}\xspace}
\newcommand{\hpnc}{{\sc HP-w/o-C}\xspace}
\newcommand{\tfobna}{\textsc{T5-O}\xspace}
\newcommand{\tfob}[1]{\textsc{T5-O-#1}\xspace}
\newcommand{\tfcb}{\textsc{T5-C}\xspace}
\newcommand{\oracle}{\textsc{Oracle}\xspace}
\newcommand{\qonly}{\textsc{Question}\xspace}

\newcommand{\nlp}[1]{\texttt{\small #1}}
\newcommand{\cmark}{\ding{51}}
\newcommand{\xmark}{\ding{55}}

\title{ASQA: Factoid Questions Meet Long-Form Answers}

\author{
    \bf Ivan Stelmakh$^{1}$\thanks{~~Work done during an internship at Google Research. Corresponding Author (\href{mailto:stiv@cs.cmu.edu}{\texttt{stiv@cs.cmu.edu}}).}
        \hspace{0.3cm}
    \bf Yi Luan$^{3}$ \\
    \bf Bhuwan Dhingra$^{2,3}$ \hspace{0.3cm}
    \bf Ming-Wei Chang$^{3}$ \\
    $^1$Carnegie Mellon University \hspace{0.3cm}
    $^2$Duke University \hspace{0.3cm} 
    $^3$Google Research \\
    }

\begin{document}
\maketitle
\begin{abstract}
\input{sections/0abstract}

\end{abstract}

\section{Introduction}
\label{section:intro}
\input{sections/1intro}

\section{Related Work}
\label{section:related}
\input{sections/2related}

\section{ASQA Task and Data}
\label{section:preliminaries}
\input{sections/3preliminaries}


\input{sections/4task}

\input{sections/5experimental_setup}

\input{sections/6experiments}

\input{sections/7analysis}

\section{Conclusion}
\label{section:conclusion}
\input{sections/8conclusion}


\bibliography{anthology,bibtex}

\begin{appendix}

\newpage

\noindent {\Large{\textbf{Appendix}}}
\input{sections/9appendix}

\end{appendix}

\end{document}

%% file: sections/0abstract.tex
An abundance of datasets and availability of reliable evaluation metrics have resulted in strong progress in \emph{factoid question answering} (QA). This progress, however, does not easily transfer to the task of \emph{long-form QA}, where the goal is to answer questions that require in-depth explanations. The hurdles include (i) a lack of high-quality data, and (ii) the absence of a well-defined notion of the answer's quality. In this work, we address these problems by (i) releasing a novel dataset and a task that we call \asqa (Answer Summaries for Questions which are Ambiguous); and (ii) proposing a reliable metric for measuring performance on \asqa. Our task focuses on factoid questions that are ambiguous, that is, have different correct answers depending on interpretation. Answers to ambiguous questions should synthesize factual information from multiple sources into a long-form summary that resolves the ambiguity. In contrast to existing long-form QA tasks (such as \eli{}), \asqa{} admits a clear notion of correctness: a user faced with a good summary should be able to answer different interpretations of the original ambiguous question. We use this notion of correctness to define an automated metric of performance for \asqa{}. Our analysis demonstrates an agreement between this metric and human judgments, and reveals a considerable gap between human performance and strong baselines.

%% file: sections/1intro.tex
In the last few years, the  factoid question answering (QA) task---extracting short answers to  \emph{factoid} questions---has witnessed  significant progress~\citep{lee-etal-2019-latent,Guu20realm,karpukhin-etal-2020-dense, lewis20augmented,izacard-grave-2021-leveraging}. The progress was achieved in large part thanks to (i) the availability of high-quality datasets~\citep{voorhees00building,joshi-etal-2017-triviaqa,yang-etal-2018-hotpotqa, abujabal-etal-2019-comqa, kwiatkowski-etal-2019-natural}, and (ii) a well-defined notion of correctness. A key challenge for ongoing research now lies in long-form question answering where the goal is to generate detailed explanations in response to  questions that require elaborate and in-depth answers. 

There is much less data available for the task of long-form QA. One of the primary data sources is the \eli{} dataset~\citep{fan-etal-2019-eli5} that pairs open-ended questions with paragraph-long answers written by users of the ``Explain Like I’m Five'' Reddit forum. However, questions in \eli{} are very general (e.g., \nlp{``How can different animals perceive different colors?''}) and can be answered in myriad different ways, making it hard to define objective criteria for a good answer. As a result, \citet{krishna-etal-2021-hurdles} identify several hurdles in using this data towards meaningful modeling progress, including a lack of reliable evaluation metrics.

\input{figures/ASQA_intro}

In this work, we address the lack of data sources and unreliability of evaluations by constructing a \emph{long-form QA dataset for factoid questions}. Our paper is motivated by the work of~\citet{min-etal-2020-ambigqa} who observe that more than half of the factoid questions that occur naturally are \emph{ambiguous}. For example, a seemingly simple question: \nlp{``Who was the ruler of France in 1830?"} is ambiguous because there were two rulers of France in 1830. \citet{min-etal-2020-ambigqa} collected the \ambig{} dataset that connects ambiguous factoid questions with  \emph{disambiguations}: pairs of disambiguated questions and unique short answers to these questions (see example on the right side of Figure~\ref{fig:example}).

We note, however, that ambiguous questions often arise when a user lacks background knowledge about \emph{why} there might be multiple answers to their question, and \emph{how} those answers relate to each other. Thus, the list of {disambiguations} may not be satisfactory for the user. For example, the fact that in 1830 the ruler of France \emph{changed due to the revolution} is highly salient but is not captured in the \ambig{} disambiguations.

In this paper, we argue the importance of generating {long-form} answers to ambiguous factoid questions. In that, we present \asqa{} (\textbf{A}nswer \textbf{S}ummaries for \textbf{Q}uestions which are \textbf{A}mbiguous)---a novel dataset that pairs each ambiguous question from \ambig{} with a crowdsourced long-form answer.\footnote{Data, evaluation scripts, and other supplementary materials are provided on the project's GitHub repository: \url{https://github.com/google-research/language/tree/master/language/asqa}} The answers we collect aim to (i) explain the source of ambiguity in the question, and (ii) connect all the valid short answers into a {coherent passage}.  An example \asqa{} instance is shown in Figure~\ref{fig:example}.

The main feature of \asqa{} is a combination of (i) \emph{a well-defined notion of correctness} pertinent to factoid QA and (ii) \emph{the complexity} of long-form QA. First, observe that a good answer to an ambiguous question should be sufficient for the user to answer different interpretations of the question. This observation induces a notion of correctness that is conceptually similar to the conventional accuracy in factoid QA. Second, to answer an ambiguous question, a system needs to retrieve a diverse set of documents that talk about different interpretations of the question and synthesize this information into a coherent summary. Thus, the key challenges of long-form QA---precise retrieval and high-quality summarization---are present in \asqa{}. 

{\paragraph{Contributions} Overall, our work makes several contributions:%
\begin{itemize}[topsep=5pt, itemsep=0pt, leftmargin=*]%
    \item First, we carefully develop a crowdsourcing pipeline and collect \asqa{}---a dataset of high-quality long-form answers to 6,316 ambiguous factoid questions.
    
    \item Second, we design principled evaluation procedures for \asqa: (i) we propose a novel automated evaluation metric (\ovscore) that combines the correctness aspect of factoid QA and the fluency aspect of long-form QA; (ii) we develop and release a convenient interface for human evaluations; (iii) we conduct a small-scale human study that shows a high agreement between our automated metric \ovscore{} and human judgments.
    
    \item Third, we establish strong baselines for our task by combining joint passage retrieval \cite{min-etal-2021-joint} and T5-large \cite{raffel19t5}. Our extensive evaluations demonstrate that there is a large gap between the baselines and human performance. Additionally, we highlight areas of improvement for future research on \asqa.
\end{itemize}
}

%% file: figures/ASQA_intro.tex
\begin{figure*}[t]
    \centering
    \includegraphics[width=\textwidth, keepaspectratio]{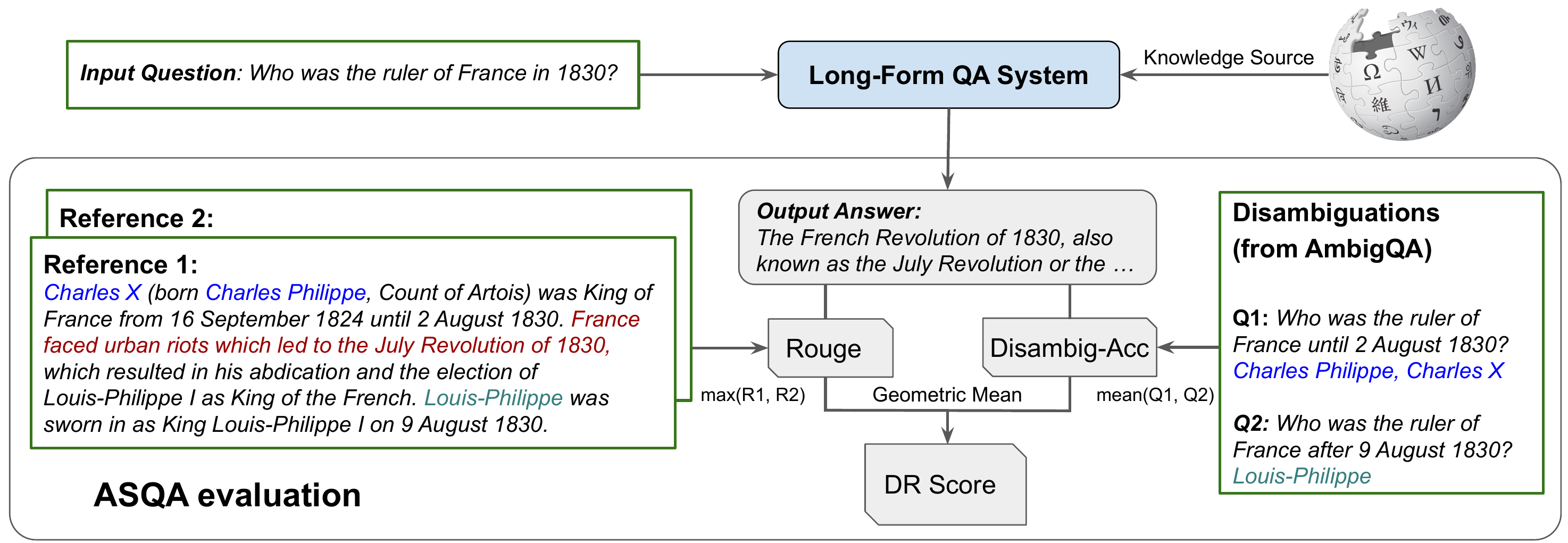}
    \caption{\asqa{} is an open-domain long-form QA dataset that focuses on answering ambiguous factoid questions.
    Input questions are sourced from \ambig{} \cite{min-etal-2020-ambigqa}. Long-form answers must be sufficient to answer disambiguated questions from \ambig{} (short answers are marked in \textcolor{blue}{\textit{blue}} and \textcolor{teal}{\textit{green}}), and should introduce additional knowledge from Wikipedia (highlighted in \textcolor{purple}{\textit{red}}) to resolve ambiguity and clarify the relationship between different short answers. The DR score we propose combines ROUGE and Disambiguation-accuracy (that is, correctness) metrics, overcoming the issues with long-form QA evaluation outlined by~\citet{krishna-etal-2021-hurdles}.
}
    \label{fig:example}
\end{figure*}

%% file: sections/2related.tex
In this section, we describe relevant works that propose 
new tasks, datasets, and methods for QA and summarization problems.

\paragraph{Extractive QA}
Much of the existing work on question answering, including
\textit{reading comprehension}~\citep{rajpurkar-etal-2016-squad, rajpurkar-etal-2018-know, trischler-etal-2017-newsqa, yang-etal-2018-hotpotqa},
\textit{open-domain QA}~\citep{kwiatkowski-etal-2019-natural, joshi-etal-2017-triviaqa}
and \textit{dialog-based QA}~\citep{choi-etal-2018-quac},
assumes that questions have unique answers. \citet{min-etal-2020-ambigqa} relax this assumption and propose a task that aims at identifying all possible short answers to the ambiguous subset of the open-domain version of the \nq{} dataset, denoted \nqopen{} \citep{kwiatkowski-etal-2019-natural, lee-etal-2019-latent}. The \ambig{} dataset constructed by~\citet{min-etal-2020-ambigqa} serves as a building block of the present work and we provide more details on this dataset in Section~\ref{section:preliminaries}.
Another related effort is the {\sc ConditionalQA} task \cite{sun2021conditionalqa}
that requires systems to identify \textit{conditions} under which the extracted answers are valid.
Unlike the \asqa task, the answers in {\sc ConditionalQA} come from a document provided in advance and do not need to be summarized into a single response.

\paragraph{Generative QA} Extractive models achieve good results when the answer to the question is readily available on the web. However, in many settings, including ambiguous factoid questions, a system needs to combine information from many (unknown) sources to present the answer to the user in a convenient way. Hence, in this work, we focus on the \emph{generative QA} setting where a model needs to generate a textual answer rather than extract it.

Datasets for generative QA include \narrative~\cite{kocisky-etal-2018-narrativeqa}
and \coqa~\cite{reddy-etal-2019-coqa}, but the average answer length in these datasets is small: 4.7 and 2.7 tokens, respectively. The MS MARCO Natural Language Generation (\msmarco{}) dataset by~\citet{nguyen2016ms} combines both extractive and generative tasks and contains slightly longer human-generated answers (usually, a sentence-long) that can be read by a smart assistant. 
\citet{fan-etal-2019-eli5} proposed a more challenging task of answering open-ended (e.g., ``why?'') questions. They scraped the \textit{``Explain Like I'm Five''} Reddit forum and released a dataset of $\sim\!\!272\text{K}$ questions, where each question is supplied with several paragraph-long answers generated by the Reddit users.
We overview the differences between \asqa{}, \eli{} and \msmarco{} in Section~\ref{sec:final-dataset}.

Recently, large language models such as GPT-3 \citep{brown2020gpt3}
have been successfully applied to the task of long-form QA using the \eli{} dataset \cite{nakano2021webgpt}. For this, a two-step human-in-the-loop approach was involved: first, demonstrations of annotators navigating the web to write answers were collected; second, a reward model~\citep{stiennon2020learning} was trained by manual pairwise comparisons of answers. In \asqa{}, relevant passages for the answer are already provided by the
annotators and we show that the proposed \ovscore~score correlates well with
the human judgment of answer quality.
Using this automated metric in place of the reward model in the approach
of \citet{nakano2021webgpt} is a potential direction for future work.

\paragraph{Summarization} 
Given a set of documents relevant to the question
(either ground truth or obtained using retrieval)
the problem of generating a long-form answer
reduces to query-based multi-document summarization.
A small-scale dataset for this task was introduced as part of
the {\sc DUC} tasks~\citep{dang2005overview}.
Recent work on building large-scale datasets has instead
focused either on query-based summarization from a single
document~\citep{nema-etal-2017-diversity,zhong-etal-2021-qmsum}
or on multi-document summarization without queries~\citep{liu2018generating,fabbri-etal-2019-multi}. In addition to the QA task, the \asqa dataset is suitable for the evaluation of systems' accuracy
in the summarization setting, where the ground-truth
passages containing the relevant information are assumed to
be given.

\paragraph{QA-Based Evaluation}
Prior work has looked at using question answering techniques to evaluate factual consistency in summarization~\citep{wang-etal-2020-asking,durmus-etal-2020-feqa} and dialogue \citep{honovich-etal-2021-q2}. These works automatically generate questions from the system-produced text and search for answers in some reference text (e.g., the input being summarized) to evaluate the quality of the output. Instead, to evaluate generated long-form answers to ambiguous questions, in \asqa we use questions created by \ambig{} annotators.

%% file: sections/3preliminaries.tex
In this section, we introduce the \asqa{} task and the underlying data-collection process. The \asqa{} task is illustrated in Figure~\ref{fig:example}. The goal of the task is to write a comprehensive paragraph-long answer $\panswer$ to a given ambiguous question $\question$.

\input{figures/ASQA_interface}

\paragraph{Source Data}
We build \asqa{} on top of the subset of ambiguous questions identified in the \ambig{} dataset. Out of a total of 14,042 \ambig{} questions, 7,207 are identified as ambiguous by at least one \ambig{} annotator. Each of these ambiguous questions $\question$ is paired with a list of $\numdis$ \textit{disambiguations} $\{(\dq_i, \da_i)\}_{i=1}^{\numdis}$, where $\dq_i$ denotes a disambiguated question and $\da_i$ denotes a unique \emph{short answer} to $\dq_i$. The number of disambiguations ranges from 2 to 46 per ambiguous question. To ensure that it is feasible to put all this information into a coherent story, we remove 417 questions with more than six disambiguations from consideration, thereby focusing on 6,790 \ambig{} instances that we use as a starting point for building our task.

\subsection{ASQA Annotation Objectives}
\label{section:objectives}

At a high level, the goal of the annotation process is to obtain high-quality long answers to ambiguous questions. We begin with a formulation of criteria for what counts as a good long answer to an ambiguous question:

\begin{itemize}[topsep=5pt, itemsep=0pt, leftmargin=*]
    \item \textbf{Completeness} The long answer should contain
    all valid short answers $\da_1, \ldots, \da_{\numdis}$ to the disambiguated questions $\dq_1, \ldots, \dq_{\numdis}$ in an appropriate context.
    
    \item \textbf{Comprehensiveness} The long answer should provide enough details for the user to (i) understand the source of ambiguity in the original question and (ii) understand the relationship between different short answers.
    
    \item \textbf{Fluency} The long answer should be coherent and fluent.
    
    \item \textbf{Attributability} The long answer should be grounded in an underlying source of information (in our case, Wikipedia).
\end{itemize}

\subsection{ASQA Annotation Process}
\label{section:annotation}

To ensure that annotations satisfy the aforementioned objectives, we develop a custom annotation interface (Figure~\ref{fig:screenshot}) and recruit native English speakers to perform our task. We then collect long-form answers for each target instance of \ambig{} using a commercial crowdsourcing platform  where it is possible to interact with the annotators on an ongoing basis. Let us now discuss the key components of our annotation pipeline.

\paragraph{Input to Annotators}   The left side of Figure~\ref{fig:screenshot} illustrates the input to our annotation procedure. Annotators are given relevant aspects of the target  \ambig{} instance: the ambiguous question $\question$, list of disambiguations $\{(\dq_i, \da_i)\}_{i=1}^{\numdis}$, and the Wikipedia pages $\wiki$ visited by \ambig{} annotators. Additionally, to help annotators understand the context behind the disambiguations without reading full Wikipedia articles, for each disambiguation $i$ we provide a (possibly empty) Wikipedia passage $\context_i$ with information relevant to the disambiguation. Details on the procedure used to find these context passages $\{\context_i\}_{i=1}^n$ are given in Appendix~\ref{appendix:annotation}.

\paragraph{Output of Annotation} The key output of annotation is a long-form answer $\gtanswer$ to a given ambiguous question $\question$. Additional elements of the output are introduced to facilitate the requirement of attributability. In that, we require annotators to provide the source Wikipedia passage $\knwldg$ for each piece of additional information they bring to their answer. Our interface has designated fields for additional knowledge (see Figure~\ref{fig:screenshot}) and annotators can add as many of these fields as they need to include any number $m$ of evidence passages $\{\knwldg_j\}_{j=1}^{\numknwldg}$.

\paragraph{Instructions and Training} The instructions for our task are written along the lines of the four criteria we discussed in Section~\ref{section:objectives} and are provided in supplementary materials. In addition to the detailed instructions, we carefully design the training procedure to minimize the amount of noise in the annotations. In that, before being accepted to the main task, annotators go through the following three-step training procedure:

\begin{enumerate}[topsep=5pt, itemsep=0pt, leftmargin=*]
    \item \textit{Self-study session} First, we give annotators a short version of the instructions. They study them on their own and then annotate three sample questions.
    
    \item \textit{In-person session} Following the self-study session, we have an online video session in which we walk annotators through the full version of the instructions and discuss mistakes made in the self-study annotations.
    
    \item \textit{Exam session} Finally, annotators complete a five-question exam. We manually evaluate all the exam answers and share personal feedback with annotators.
\end{enumerate}

In total, 27 annotators went through our training procedure and all of them were eventually accepted to work full-time on the main task. We note that the quality of answers in the self-study session was very diverse with some annotators making critical mistakes (e.g., not covering some of the disambiguations). However, the in-person session proved to be efficient in helping annotators to understand the requirements, leading to exam answers of consistently high quality.

\smallskip

\noindent Having introduced the main aspects of the annotation procedure, we now proceed to the discussion of collected data. Additional details on the annotation procedure are given in Appendix~\ref{appendix:annotation}.

\subsection{ASQA Dataset}
\label{sec:final-dataset}
By following the procedure outlined above, we annotated train, dev, and test splits of the \ambig{} dataset. Each question in the train split was annotated by a single annotator while the dev and test splits have two annotations per question.

For 474 questions, our annotators raised concerns regarding the validity of the \ambig{} disambiguations. Not all of these concerns necessarily indicate errors in the \ambig{} dataset as some of them could be due to misinterpretation on the annotators' side. Nevertheless, to maintain data fidelity, we exclude the corresponding instances from the resulting dataset. Table~\ref{table:datastat} displays the final breakdown of the \asqa{} dataset.  

\begin{table}[t]
\begin{center}
\begin{small}
\begin{sc}
\begin{tabular}{lrr}
\toprule
Split & \# Questions & \# Annotations \\
\midrule
Train   & 4,353   & 1  \\
Dev     & 948     & 2  \\
Test    & 1,015   & 2  \\
\bottomrule
\end{tabular}
\end{sc}
\end{small}
\end{center}
\vskip -0.1in
\caption{Summary statistics of the \asqa{} dataset.
}
\label{table:datastat}
\end{table}

\input{tables/datasets_comparison}

Table~\ref{table:datasets} compares the \asqa{} dataset to other open-domain QA datasets: \eli{}, \msmarco{} (MS MARCO Natural Language Generation task), \ambig{}, and \nqopen{}. We make two main observations from the table. First, \asqa{} requires answers on the longer side of the spectrum with an average length of 64.8 (vs. 103.0 for \eli{} and 14.6 for \msmarco{}). Second, \asqa{} is the only dataset that admits evaluations in terms of both (i) \textsc{ROUGE} metrics that are typically used in long-form QA and (ii) accuracy metric that is typically used in factoid QA. 

These features make \asqa{} an appealing dataset for research on question answering as it enables researchers to work on long-form QA while retaining the benefits of reliable objective evaluation typical for factoid QA. More details on the evaluation metrics for \asqa are given in Section~\ref{section:task}.

\paragraph{Additional Comparison to ELI5} \eli{} is the closest existing long-form QA dataset. We now provide additional comparison of \asqa{} and \eli{}.

\smallskip

\noindent \textit{Support Documents} First, both \asqa{} and \eli{} supplement annotations with relevant information retrieved from Wikipedia (\asqa) or the whole Internet (\eli). For \eli{}, support documents are retrieved automatically and independently of the annotation process. The resulting documents contain, on average, 858 words. Manual analysis conducted by~\citet{fan-etal-2019-eli5} reveals that support documents are sufficient to answer 65\% of the questions and have information relevant to 92\% of the questions.

In \asqa{}, support documents are constructed as a part of the annotation process. For each annotation, the support document contains disambiguations from \ambig{}, context paragraphs, and additional knowledge provided by the corresponding annotator (see Section~\ref{section:annotation} for details). On average, support documents contain 225 words, being much shorter than those for \eli{}. By design of our annotation procedure, support documents should be sufficient to answer ambiguous questions. Indeed, we observe that 92\% of the annotations' tokens are present in the corresponding support documents.\footnote{This statistic is computed as the {\sc ROUGE1} recall score between lowercased annotations and support documents. In this work, we use {\sc rouge-score 0.0.4} python package for all {\sc ROUGE} computations.} If we exclude \ambig{} disambiguations from the support documents, their average length reduces to 172 words, but 78\% of tokens from the answers remain captured therein.  These observations demonstrate that \asqa{} satisfies the requirement of attributability formulated in Section~\ref{section:objectives}.

\smallskip

\noindent \textit{Inter-Annotator Agreement} Second, we compare the inter-annotator agreement in \eli{} and \asqa{} that we measure as the mean \lrouge F1 score between each pair of annotations for the same question. Our analysis reveals that \asqa{} has a much higher level of inter-annotator agreement: 49.6 vs. 16.9 for \eli{}. Thus, \asqa{} admits a more well-defined notion of ground truth than \eli{}. 

Note that answers in \eli{} are written by Reddit users. Thus, they are inherently subjective and are not supposed to follow any predefined criteria. The diversity and subjectiveness could make human evaluation of the \eli{} answers challenging. In contrast, \asqa{} annotators follow common annotation guidelines and undergo a thorough training procedure, thereby aiming at generating answers that satisfy a set of well-defined criteria for human evaluation (Section~\ref{section:objectives}).  

As a note of caution, we remark that the high inter-annotator agreement in \asqa{} is contingent upon the high inter-annotator agreement in the \ambig{} dataset. Indeed, \ambig{} disambiguations serve as a shared source of information between two \asqa{} annotators working on the same instance, potentially inflating the level of agreement. That said,~\citet{min-etal-2020-ambigqa} observe that human annotators have a decent level of agreement in constructing the disambiguations, thereby supporting the observation that \asqa{} is more objective than \eli{}. 

\smallskip

Overall, compared to other datasets, \asqa{} has some novel features that may be useful for future QA research. Its benefits, however, come at the cost of a much smaller sample size than that of \msmarco{} and \eli{}. Thus,  we believe \msmarco{} and \eli{} may be useful counterparts for \asqa{} as they can be used for pre-training  (that said, we leave this exploration to future work).

%% file: figures/ASQA_interface.tex
\begin{figure*}[t]
    \centering
    \includegraphics[width=14.75cm]{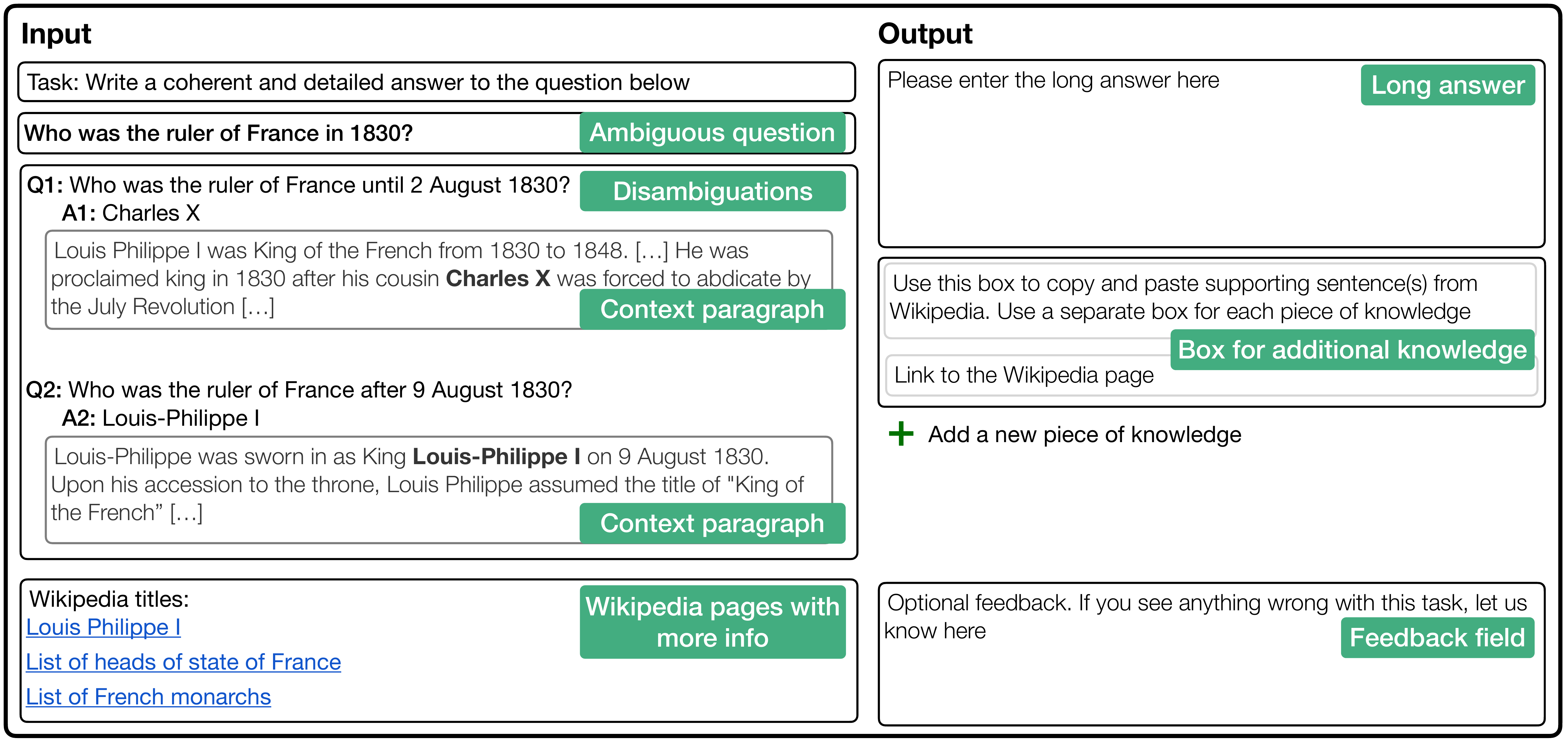}
    \caption{Schematic representation of the annotation interface.
    }
    \label{fig:screenshot}
\end{figure*}

%% file: tables/datasets_comparison.tex
\begin{table*}[tb]
\begin{center}
\begin{small}
\begin{sc}
\begin{tabular}{p{2.2cm}rrrrcc}
\toprule
 \multirow{2}{*}{QA Task} & & &  \multicolumn{2}{c}{Dev Set Statistics} &
 \multicolumn{2}{c}{Evaluation} \\
  \cmidrule(lr){4-5} \cmidrule(lr){6-7}
  & Dataset & \#QAs& \#A per Q&  \#Words in A & ROUGE & Disambig-Acc  \\
 \midrule
 \multirow{2}{*}{Short Answer}& \nqopen{} & 91K & 1.8 & 2.2 & \xmark & \cmark{$^\dagger$} \\
& \ambig{} & 14,042 & 2.8 & 2.4 & \xmark & \cmark \\
\midrule
\multirow{3}{*}{Long Form} 
& \eli{} & 272K & 12.0 & 103.0 & \cmark & \xmark \\
& \msmarco{} & 183K & 1.7 & 14.6  & \cmark & \xmark \\
\cmidrule(lr){2-7}
&\asqa{} & 6,316 & 2.0 & 64.8 & \cmark & \cmark \\
\bottomrule
\end{tabular}
\end{sc}
\end{small}
\end{center}
\vskip -0.1in
\caption{Comparison of \asqa with existing open domain QA datasets.
\asqa is the only QA dataset that allows for both ROUGE and accuracy evaluations. $^\dagger$Standard accuracy for non-ambiguous questions.
} 
\label{table:datasets}
\end{table*} 

%% file: sections/4task.tex
\section{ASQA Metrics}
\label{section:task}

In this section, we introduce metrics that we propose to evaluate performance on the \asqa{} task.

\subsection{Automated Evaluation}

We evaluate performance on the \asqa
task along the following two aspects. 

\paragraph{ROUGE} Following the conventional approach of measuring the quality of the generated text, including long-form answers~\citep{fan-etal-2019-eli5}, we report the \lrouge score~\citep{lin04rouge} in a multi-reference setup.\footnote{We lowercase candidate and reference summaries and compute the {\sc rougeLsum} F1 score with stemmer on.} Specifically, given that each example in the development and test sets is annotated by two annotators, we compare predictions against both answers and take the \textit{maximum} of these two scores to be the score of the prediction.

\paragraph{Disambiguation Metrics} As discussed in Section~\ref{section:objectives}, a good long-form answer to an ambiguous question should contain short answers to all disambiguated questions as well as the context necessary to understand the source of ambiguity and the relationship between the short answers. However, \lrouge{} is not well suited for evaluating these aspects as it may fail to distinguish between two fluent and stylistically similar answers which provide considerably different information. With this motivation, we complement \lrouge{} with two metrics that are specifically designed to capture the \emph{completeness} and \emph{comprehensiveness} aspects of our task:
\begin{itemize}[topsep=5pt, itemsep=0pt, leftmargin=*]
    \item \strem~(String Exact Match)  The fraction of disambiguations for which the corresponding short answer is present in the long answer (exact match). The fraction is first computed within each question and then averaged across all questions.

    \item \qaf  As a more principled approach toward measuring the informativeness of a long answer, we follow the reading comprehension literature~\citep{rajpurkar-etal-2016-squad, rajpurkar-etal-2018-know}. In that, we use the Roberta model~\citep{liu19roberta} pretrained on \squad{} to evaluate the fraction of disambiguated questions that can be answered from the predicted long answers.\footnote{We use Huggingface training and evaluation scripts~\citep{wolf-etal-2020-transformers}.} 
    Specifically, for each disambiguation $(\dq_i^{(k)}, \da_i^{(k)})$ in the $k$-th example, we apply the \squad~model on the generated long-form answer $\panswer^{(k)}$ to predict short answer $\pda_i^{(k)}$ to question $\dq_i^{(k)}$. Let $\fone$ denote a function that computes the token-level F1 score between the predicted short answer $\pda_i^{(k)}$ 
    and the ground truth short answer $\da_i^{(k)}$ after
    normalizing answer strings in the manner done for \squad{} evaluations. Then the \qaf score is given by:
    \begin{align*}
    \text{\qaf} = \frac{1}{\samplesize} \sum_k \frac{1}{\numdis^{(k)}}\sum_{i} \fone (\pda_i^{(k)}, \da_i^{(k)}),
    \end{align*}
    where $\samplesize$ indicates the total number of \asqa{} instances in the development or test split, and $\numdis^{(k)}$ indicates the number of disambiguations for the $k$-th instance.
\end{itemize}

As a note of caution, we remark that the above disambiguation metrics only measure the \textit{recall} of the required information in the long answers. In cases where the long answer contains additional information that pertains to incorrect disambiguations (e.g., an additional hallucinated disambiguation), unnecessary extra information will be penalized by the \lrouge metric. Moreover, even for disambiguation metrics, it is likely that the accuracy of the Roberta model will decrease due to the presence of distractors, thereby effectively penalizing a low precision.

\paragraph{Overall: \ovscore{} Score}
Both \lrouge{} and disambiguation metrics are crucial for our task. Hence, we propose an overall \ovscore{} (Disambiguation-Rouge) score that combines the two metrics as follows:
\begin{align*}
    \text{\ovscore} = \sqrt{\text{\qaf} \times \text{\lrouge}}.
\end{align*}
We choose the geometric mean for aggregation to penalize methods that maximize one metric at the cost of a significant decrease in the other. Note that \strem{} and \qaf{} aim at measuring the same aspect so we include only one of these metrics in the \ovscore{} score. 
\subsection{Human Evaluation}
\label{sec:human_eval}
Despite our automated metrics being aimed at the crucial aspects of the \asqa{} task, they are not guaranteed to closely approximate the judgment of humans, whose satisfaction is an overarching goal of a QA system. Therefore, we design and release an interface for human evaluations for the \asqa{} task, which operates with the following metrics:

\begin{enumerate}[topsep=5pt, itemsep=0pt, leftmargin=*, label=(\roman*)]
    \item \emph{Disambiguation Accuracy} For each long-form answer, we ask human annotators to verify whether each disambiguated question from the \ambig{} dataset can be correctly answered using the provided information. We then report the average number of disambiguations that are captured in the long-form answers (\acc).

    \item \emph{Pairwise Comparisons} Next, we propose a pairwise evaluation scheme where annotators need to compare two long-form answers to the same question (either from two different models or one from a model and one reference). Specifically, we ask annotators to choose the better answer in terms of each of the three criteria: Comprehensiveness (\comp), Fluency (\flue), and Human Overall impression (\hovscore). In each pairwise comparison, an answer is given one point for victory and half a point for a tie. We then convert model scores into percentages by dividing the total number of points a model receives by the number of pairwise comparisons it is evaluated in and multiplying this ratio by 100.
\end{enumerate}

In our experiments (Section~\ref{sec:results}), both disambiguation accuracy evaluations and pairwise comparisons are conducted in a blind fashion with annotators being unaware of the source of the answers they evaluate. The scripts for recreating this annotation interface are provided in supplementary materials.

%% file: sections/5experimental_setup.tex
\section{Experimental setup}
\label{section:experiment_setup}

We now describe the baseline models and reference human answers used for comparison in our experiments.
\subsection{Models}
\label{section:models}
We include the following models to comparison.

\paragraph{Na\"ive} To establish a lower bound on the task, we use the \qonly model that repeats the ambiguous question eight times to make the length comparable to the outputs of other systems. 

\paragraph{Retrieval-Only} The next pair of models retrieve a Wikipedia passage in response to a given question:
\begin{itemize}[topsep=5pt, itemsep=0pt, leftmargin=*]
    \item \dprk{1}. \dpr~\citep{karpukhin-etal-2020-dense} is a BERT-based dual encoder trained on NQ.
    
    \item \jprk{1}. \jpr~\citep{min-etal-2021-joint} trains a reranker on top of \dpr{} for questions with multiple answers in \ambig{}. The \jpr{} model is a state of the art on \ambig{} at the time of writing.
\end{itemize}

\paragraph{Generative} We also evaluate T5-based generative models~\citep{raffel19t5} in two regimes:
\begin{itemize}[topsep=5pt, itemsep=0pt, leftmargin=*]
    \item T5 Closed Book (\tfcb).  In the closed book setup, we train T5 to answer ambiguous questions without providing any additional passages from Wikipedia. The model only relies on its pretrained knowledge to answer the question \cite{roberts-etal-2020-much}.
    
    \item T5 Open Book (\tfobna). In the open book setup, the T5 model is additionally provided with contexts paragraphs retrieved by \jpr. We vary the number of top-$K$ retrieved paragraphs used as input to T5, denoting the corresponding model as \tfob{K}.
\end{itemize}
In both cases, we use the T5-large model.

\paragraph{Oracle} To investigate the headroom in retrieval systems, we introduce the \oracle{} system: T5-large provided with support documents (that is, the knowledge required to answer the question). Specifically, the input to \oracle{} includes all the disambiguations $\{(\dq_i, \da_i)\}_{i=1}^{\numdis}$ and contexts $\{\context_i\}_{i=1}^{\numdis}$ shown to the annotators (left half of Figure~\ref{fig:screenshot}),
as well as the additional knowledge pieces $\{\knwldg_j\}_{j=1}^{\numknwldg}$ identified by one of the two annotators (the one with the longest answer).
This system can be thought of as a generative model that has access to a perfect retriever. In evaluations, we compute \lrouge{} by comparing the answer predicted by \oracle{} against the answer of the annotator whose additional knowledge pieces \emph{were not} in the input of \oracle{} (instead of the usual comparison against two references). 

\medskip

Appendix~\ref{appendix:model} provides more details on the modeling aspects of our evaluations.

\subsection{Human-Generated Answers} 
\label{section:human_answers}

Finally, we evaluate two sets of human-generated answers:

\begin{itemize}[topsep=5pt, itemsep=0pt, leftmargin=*]
    \item Human performance with context (\hpwc). First, we use reference \asqa{} answers in our comparisons. Recall that \asqa{} annotators were provided with common context: disambiguations from \ambig{} $\{(\dq_i, \da_i)\}_{i=1}^{\numdis}$ and context paragraphs we retrieved $\{\context_i\}_{i=1}^{\numdis}$. Therefore, we consider performance in this setup as an upper bound on the human performance. In evaluations of \lrouge{}, we compute the score of \hpwc{} by comparing the answers from two annotators against each other (instead of the usual comparison against two references).
    
    \item Human performance without context (\hpnc). To establish a conservative lower bound on human performance, we additionally annotate 200 questions from the \asqa{} dev set (one annotation per question) in the ``no context'' regime. Specifically, annotators in this regime are only given ambiguous questions as input (no disambiguations and context paragraphs) and need to search for disambiguations and the required additional information on their own. We do not specifically train annotators for these additional annotations and just ask them to provide answers of the same quality as in the main task.
\end{itemize}

%% file: sections/6experiments.tex
\section{Results}
\label{sec:results}

\def\arrvline{\hfil\kern\arraycolsep\vline\kern-\arraycolsep\hfilneg}

\input{tables/results}

We evaluate all models introduced above in the automated evaluations. Additionally, we conduct a small-scale human study involving a subset of models to provide some verification of the automated evaluation results. Specifically, our human study involves four model outputs (\jprk{1}, \tfcb, \tfob{1}, \tfob{5}) and two sets of human-generated answers (\hpnc, \hpwc) that are juxtaposed on a subset of 45 randomly chosen questions from the development set of \asqa{}. For each of the questions, six target answers are split into three pairs and pairwise comparisons are conducted by authors of this paper in a blind manner.

Table~\ref{table:main_results} and Table~\ref{table:human_eval} display the results of automated and human evaluations, respectively. Our main observations are as follows.

\paragraph{Importance of Retrieval}  Models that take the output of a retrieval system perform much stronger than the closed-book model. The T5 open-book setups (\tfob{1/3/5}) all outperform the T5 closed-book setup (\tfcb) on all of the automated metrics as well as in the human evaluation. In that, \tfob{1} outperforms \tfcb by 20.0 points on human evaluation (\hovscore) and by 13.5 points on \ovscore.
\tfob{5} outperforms \tfcb by 15.6 points on \hovscore{} and by 18.0 points on \ovscore. Additionally, when the oracle passages are provided to T5, the \ovscore score is more than three times that of \tfcb. Thus, we conclude that, in contrast to \eli, strong performance on our task is contingent upon having a high-quality retrieval model. We further analyze the importance of retrieval on \asqa in Section~\ref{section:analysis_retrieval}.

\paragraph{Importance of Summarization} Retrieval is very important for \asqa, but just using the top retrieved passage from a strong system (\jprk{1}) is not sufficient. Indeed, even though the \strem and \qaf metrics of \jprk{1} are considerably higher than these of \tfob{1} (by 11.4 and 4.6 points, respectively), the human overall impression score \hovscore{} is similar across these models. This discrepancy is observed because the disambiguation metrics do not evaluate the conciseness of the answers, and the advantage of \textsc{JPR@1} on these metrics is gained at the cost of the increased answer length (196.8 words). In contrast, T5 models tend to generate shorter answers whose length is much closer to the average length of human references (65 words). Hence, in addition to including the correct information, answers in \asqa must be concise which highlights the importance of summarization.

\input{tables/human_eval}

\paragraph{Correlation with Human Judgments} Table~\ref{table:correlation} reports Pearson correlations between different automated metrics and the human judgments, enabling us to study the validity of the automated metrics. 

First, we observe that \qaf{} is much better correlated with human evaluations than \lrouge{}. That said, we note that \lrouge{} is an important metric as it enforces conciseness of the answers. Additionally, \lrouge{} may become better correlated with human judgments when models improve on the  \qaf{} metric.

Second, observe that \qaf{} scores (Table ~\ref{table:main_results}) underestimate the human evaluations of \acc{} (Table~\ref{table:human_eval}). This discrepancy is likely due to: (i) a distribution shift between \asqa{} and \squad{}; and (ii) the presence of distracting answers from the other disambiguated questions in the long answers, which are known to degrade QA models' accuracy \cite{jia-liang-2017-adversarial}. However, almost perfect correlation between \qaf and \acc{} (99.4) implies that this discrepancy does not impact the ordering of the answers, thereby enabling us to meaningfully evaluate the relative differences in performance. Additionally, the presence of strong distractors ensures that the \qaf{} metric cannot be easily gamed by mentioning all the short answers without appropriate context.

Finally, we note that the \ovscore{} score has the highest correlation with the overall human judgment \hovscore{} among all automated metrics. While the difference with \qaf{} is not statistically significant, this observation hints at the importance of combining \lrouge{} and \qaf{} in the overall metric to take a holistic view on the model performance.

\input{tables/correlation_table}

\paragraph{Remaining headroom}
Both the upper bound (61.8 \ovscore and 88.9 \hovscore) and the lower bound (42.3 \ovscore and 74.4 \hovscore) on human performance do significantly exceed the best model performance (\tfob{5} with 33.7 \ovscore and 36.7 \hovscore). Hence, there is a lot of headroom for the community to explore in \asqa, and we report some additional insights that may be helpful for future work in Section~\ref{section:analysis_retrieval}.

%% file: tables/results.tex
\begin{table*}[htb]
\begin{center}
\begin{small}
\begin{sc}
\begin{tabular}{lcccc|c}
\toprule
 & Len (wrds) & {\lrouge} & \strem & {\qaf}  & \ovscore \\
\midrule
\qonly & $71.6$ & $15.3$ & $1.2$ & $0.2$  & $1.5$\\
\midrule
\dprk{1} &  $99.9$ & $33.8$ & $30.1$ & $16.7$  & $23.7$\\
\jprk{1} & $196.8$ & $30.5$ & $45.0$  & $25.8$  & $28.1$\\
\midrule
T5 Closed Book (\tfcb) & $62.5$ &  $33.5$ & $10.3$  & $7.4$  & $15.7$\\
T5 Open Book 1 passage (\tfob{1}) & $63.0$ & $40.3$ & $33.6$ &  $21.2$  & $29.2$\\
T5 Open Book 3 passages (\tfob{3}) & $71.1$ & $42.7$ & $39.9$  & $25.1$ & $32.7$\\
T5 Open Book 5 passages (\tfob{5}) & $71.6$ & $43.0$ & $41.0$ & $26.4$  & $33.7$\\
\midrule
\midrule
T5 Open w/ Oracle Context (\oracle) & $82.6$ & $46.6^{\ast}$ & $88.7$ & $59.2$ & $52.5^{\ast}$\\
Human w/o Context (\hpnc) & $73.5$ & $45.8$ & $51.8$ &  $39.0$ & $42.3$\\
Human w/  Context (\hpwc) & $64.8$ & $49.4^{\ast}$ & $98.4$ &  $77.4$ & $61.8^{\ast}$\\
\bottomrule
\end{tabular}
\end{sc}
\end{small}
\end{center}
\vskip -0.1in
\caption{Evaluation of baselines on the dev set of the \asqa task. T5 models with passages retrieved
by \jpr{} are the best models, but there is a large gap between human performance and model performance on all metrics.
$^\ast$As explained in Section~\ref{section:experiment_setup}, for \oracle{} and \hpwc we only use one of the references to compute \lrouge.}
\label{table:main_results}
\end{table*}

%% file: tables/human_eval.tex
\begin{table}[bt]
\begin{center}
\begin{small}
\begin{sc}
\begin{tabular}{lrrrrrrrr}
\toprule
 & Acc & Comp & Flue & \hovscore \\
\midrule
JPR@1 & {36.1} & {44.4} & {42.2} & 37.8\\
T5 C & 8.4 & 35.6 & 32.2 & 21.1\\
T5 O-1 & 25.7 & 36.7 & 38.9 & {41.1} \\
T5 O-5 & 28.0 & 36.7 & 37.8 & 36.7 \\
\midrule
\hpnc & 52.7 & 60.0 & 66.7 & 74.4 \\
\hpwc & 94.3 & 86.7 & 82.2 & 88.9 \\
\bottomrule
\end{tabular}
\end{sc}
\end{small}
\end{center}
\vskip -0.1in
\caption{\label{table:human_eval} Results of human evaluations executed on a set of 45 questions from the development set of \asqa{}.  The score range is from 0 to 100 and larger values are better. All metrics are specified in Section~\ref{sec:human_eval}.}
\end{table}

%% file: tables/correlation_table.tex
\begin{table}[tb]
\begin{center}
\begin{small}
\begin{sc}
\begin{tabular}{lrrr}
\toprule
 & {\lrouge} & {\qaf} & \ovscore  \\
\midrule
\acc & 72.4 & {99.4} & 97.7 \\
\comp & 69.0 & {96.4} & 92.7 \\
\flue & 76.4 & {94.4} & {94.4}  \\
\hovscore & 81.9 & 92.9 & {95.2} \\
\bottomrule
\end{tabular}
\end{sc}
\end{small}
\end{center}
\vskip -0.1in
\caption{\label{table:correlation} Correlation between human evaluation metrics and automated metrics. Scores range from 0 to 100 and larger values indicate stronger correlations. \ovscore{} has the highest correlation with the 
overall human score \hovscore{} among all automated metrics.}
\end{table}

%% file: sections/7analysis.tex
\section{Analysis}

We now conduct additional analysis that provides insights on the \asqa{} task.

\subsection{Quantitative Analysis}
\label{section:analysis_retrieval}
\input{figures/passage_retrieval}

We begin with a quantitative analysis that investigates room for improvement in both summarization and retrieval aspects of \asqa{}. 

\paragraph{Headroom in Summarization} As shown in Figure~\ref{fig:qaf1_numpassages}, the \qaf performance of retrieval-based methods increases considerably as the number of retrieved passages increases. However, there is a big gap between T5 and \jpr{}, even though T5 takes the output passages from \jpr{} as an input. This observation indicates that T5 tends to either lose information while summarizing the passages or produce outputs that are inconsistent with its input. Moreover, the \qaf of \jprk{5} already exceeds the corresponding value for the lower bound on the human performance. Thus, progress in summarization alone may be sufficient to raise the overall level of performance on \asqa{} to this lower bound.

\paragraph{Headroom in Retrieval} Figure~\ref{fig:qaf1_numpassages} demonstrates that the best-performing retrieval system, \jprk{5},
lags behind the output of the \oracle{} model by 14.4 and the human upper bound by 32.6. Hence, improving the retrieval step for \asqa is also important. This observation is further evidenced by the fact that \ovscore of the \oracle{} model (52.5) is much higher than that of \tfob{5} (33.7).

\subsection{Qualitative Analysis}
\input{tables/analysis}

To provide further insight into the importance of the generation aspect of our task, we conduct a manual analysis of the answers generated by the T5 open-book model. Our main observation is that even if the knowledge necessary to answer an ambiguous question \textit{is successfully retrieved}, T5 often struggles to provide a high-quality answer. Table~\ref{tab:examples} demonstrates several characteristic mistakes that we identify.

\paragraph{Hallucination} The first two rows of Table~\ref{tab:examples} demonstrate examples of hallucination in the T5-generated answers. In the first example, T5 hallucinates facts about the \emph{2016 elections} (there were no elections in 2016) and about \emph{the winner of the 2017 elections} (Rick Baker did not win the elections). In the second example, T5 starts with a wrong disambiguation (dragons do not marry people) and then mixes up facts about two characters from different books (\textit{Daenerys
Targaryen} and \textit{Elizabeth/Liz Pennykettle}) into one. 

\paragraph{Question Misunderstanding} Another issue we observe in the T5-generated answers is that sometimes the answers provide a coherent story that is relevant to the question but does not answer it. This problem is illustrated in the third row of Table~\ref{tab:examples} where the question ``\emph{When was <<under God>> added to the Pledge of Allegiance?}'' is answered with a history of the Pledge of Allegiance but does not mention the target phrase (<<under God>>).

\paragraph{Repetitions} Finally, we observe a somewhat technical issue of repetitions in the generated answers,
as shown in the second row of Table~\ref{tab:examples}.

\smallskip

\noindent We conclude that summarization is an important component of ASQA which still has a lot of room to be improved in order to meet the human-level performance on the whole task.

%% file: figures/passage_retrieval.tex
\begin{figure}[tb]
    \centering
    \includegraphics[width=\linewidth, trim={0mm 0mm 0mm 0mm}, clip=on]{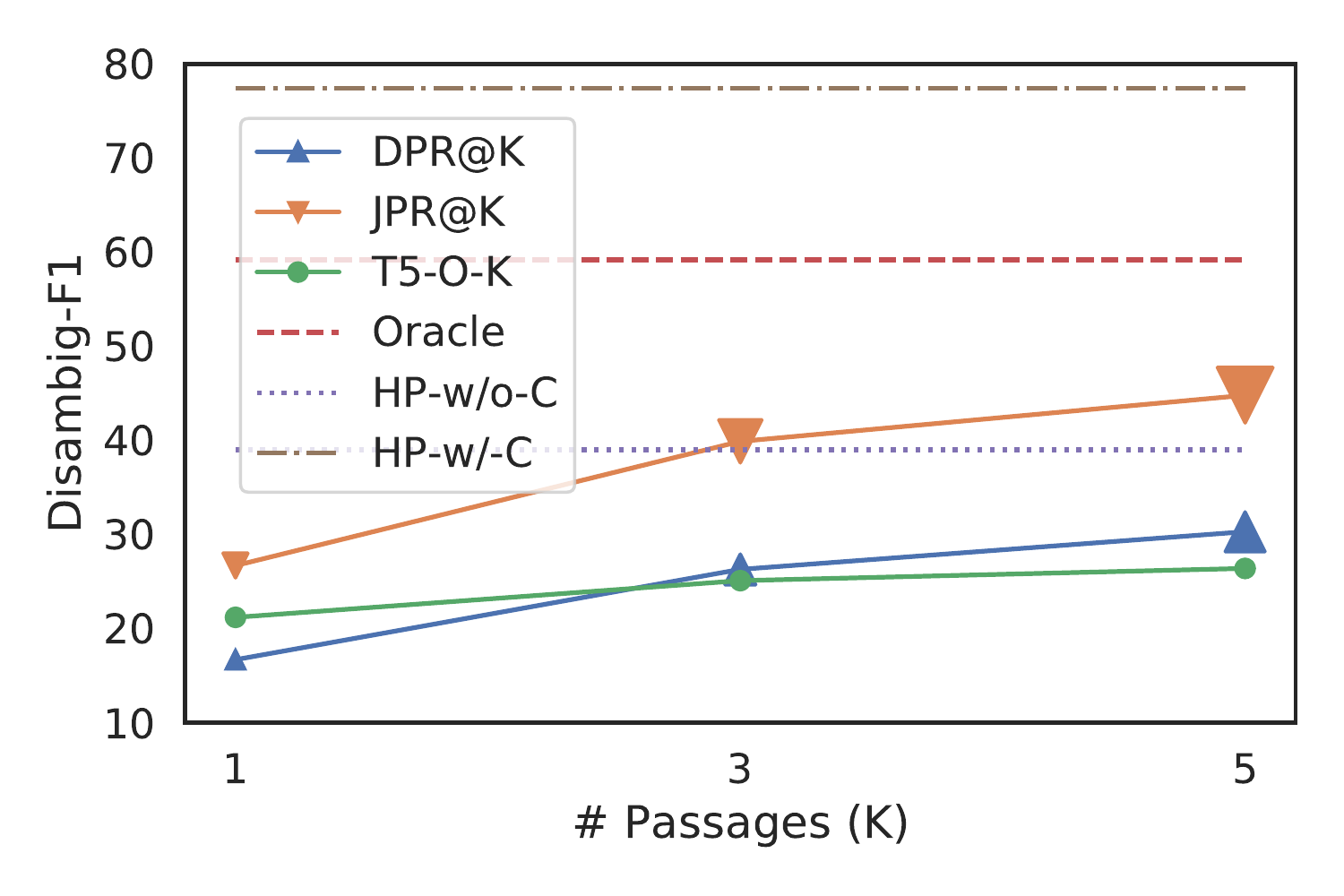}
    \caption{\qaf of different methods with a varying number of retrieved passages.
    Marker sizes are proportional to the answer lengths.
    The \tfob{K} score increases with $K$ but there is also an increasing gap between
    \tfob{K} and \jprk{K}. Passages from the latter are used as input for the former.}
    \label{fig:qaf1_numpassages}
\end{figure}

%% file: tables/analysis.tex
\begin{table*}
\centering
\footnotesize
\begin{tabular}{p{4.25cm}|p{3.5cm}|p{3.85cm}|p{2.7cm}}
\toprule
 Question and disambiguations & T5 output & JPR@5 relevant segments & Error analysis \\
\midrule
\textit{Q}: Who won the mayor race in st petersburg florida? \newline
\newline
\textit{Disambig-Q1}: Who won the 2017 mayor race in st petersburg florida? \newline
 Kriseman | Rick Kriseman
\newline
\textit{Disambig-Q2}: Who won the 2013 mayor race in st petersburg florida? \newline
Kriseman | Rick Kriseman
\newline
\textit{Disambig-Q3}: Who won the 2009 mayor race in st petersburg florida? \newline
Foster | Bill Foster &
the city of st. petersburg, florida has had several mayoral elections. \textcolor{red}{democrat rick kriseman won the 2016 mayoral election, while republican former mayor rick baker did so in the 2017 mayoral election.} kriseman defeated baker in the 2017 general election with 51.62\% of the vote.
&
\textit{2017 St. Petersburg, Florida mayoral election}: \dots Incumbent mayor Rick Kriseman, a Democrat, opted to seek re-election. His opponent in the runoff was Republican former mayor Rick Baker \dots Incumbent Rick Kriseman won the re-election\dots \newline
\textit{Rick Kriseman}:  On November 7, 2017, Kriseman was re-elected for a second term, winning the general election with 51.62\% of the vote. He defeated former Republican Mayor Rick Baker\dots & T5 hallucinates (i) the 2016 mayoral elections (there were no elections in 2016) and (ii) the winner of the 2017 mayoral elections (Kriseman won elections, but the model claims Baker to be the winner). \\ 
\midrule

\textit{Q}: Who was the mother of dragons married to? \newline
 \newline
\textit{Disambig-Q1}: Who was the mother of dragons first married to? \newline
Khal Drogo | Drogo \newline
\textit{Disambig-Q2}: Who was the mother of dragons married to second? \newline
Hizdahr zo Loraq
\newline
\textit{Disambig-Q3}: What leader of Dothraki is the mother of dragons married to? \newline
Khal Drogo \newline
\textit{Disambig-Q4}: What Ghiscari noble is the mother of dragons married to? \newline
Hizdahr zo Loraq &
\textcolor{red}{dragons are often married to multiple people in a song of ice and fire storyline. in a song of ice and fire, the mother of dragons is known as elizabeth/liz pennykettle}, a woman probably in her thirties who makes dragons out of clay and sells them at pottery fairs. she is the mother of lucy pennykettle, wife of the blinded ex-monk arthur, and landlady of david rain. elizabeth/liz pennykettle - \textcolor{blue}{a woman probably in her thirties who makes dragons out of clay and sells them at pottery fairs. she is the mother of lucy pennykettle, wife of the blinded ex-monk arthur, and landlady of david rain.}
&
\textit{A Song of Ice and Fire}:
\dots Daenerys is married off by her elder brother Viserys Targaryen to  \dots \newline
\textit{Daenerys Targaryen}:
 Viserys marries Daenerys to the powerful Dothraki warlord Khal Drogo in exchange for his military support  \dots \newline
\textit{The Last Dragon Chronicles}:
 Elizabeth/Liz Pennykettle – A woman probably in her thirties who makes dragons out of clay and sells them at pottery fairs. 
 & In the first sentence of the answer, T5 hallucinates that dragons are married to humans in the \emph{A Song of Ice and Fire} storyline. In the second sentence, T5 suddenly jumps from a discussion of \emph{A Song of Ice and Fire} to \emph{The Last Dragon Chronicles}, thereby misleading the reader. \newline

The blue text highlights repeated content.
 
 \\ 
 \midrule
 \textit{Q}: When was under god added to the pledge of alligence? \newline
\newline 
\textit{Disambig-Q1}: When was under god officially added to the pledge of alligence? \newline
June 14, 1954 \newline
\textit{Disambig-Q2}: When did Bowman popularly introduce the pledge of alligence with under god added to a society meeting in Chicago? \newline
February 12, 1948
\newline
\textit{Disambig-Q3}: When was the bill signed into law that  under god was added to the pledge of Allegiance? \newline
Flag Day | June 14, 1954 \newline
\textit{Disambig-Q4}: When was under god added to the pledge of Allegiance for the first time? \newline
February 12, 1948 \newline
\textit{Disambig-Q5}: When did the board of directors of the Knights of Columbus adopt a resolution to add under god to the pledge of Allegiance? \newline
April 30, 1951 
&
the pledge of allegiance of the united states is an expression of allegiance to the flag of the united states and the republic of the united states of america. it was originally composed by captain george thatcher balch, a union army officer during the civil war and later a teacher of patriotism in new york city schools. the form of the pledge used today was largely devised by francis bellamy in 1892, and formally adopted by congress as the pledge in 1942. the most recent alteration of its wording came on flag day, june 14, 1954. \textcolor{red}{[Missing mention of ``under god'' words]} &
\textit{Pledge of Allegiance} Louis Albert Bowman, an attorney from Illinois, was the first to suggest the addition of "under God" to the pledge \dots At a meeting on February 12, 1948, he led the society in reciting \dots  \newline
\textit{Pledge of Allegiance} In 1951, the Knights of Columbus, the world's largest Catholic fraternal service organization, also began including the words "under God" in the Pledge of Allegiance.  \newline
\textit{Pledge of Allegiance} Congress passed the necessary legislation and Eisenhower signed the bill into law on Flag Day, June 14, 1954. Eisenhower said: The phrase "under God" was incorporated into the Pledge of Allegiance on June 14, 1954, by a Joint Resolution of Congress amending § 4 of the Flag Code enacted in 1942.

& The T5 output introduces the Pledge of Allegiance and mentions some of the right dates (June 14, 1954), but does not mention that alteration on June 14, 1954, included the words ``under god'' to the Pledge.
 \\ 

\bottomrule
\end{tabular}
\caption{\label{tab:examples} Error analysis for \tfob{5}. The colored text highlights problematic parts of the T5 output.}
\vspace{-1mm}
\end{table*}

%% file: sections/8conclusion.tex
In this work, we release \asqa{}---a novel dataset for long-form question answering. In contrast to existing datasets for long-form QA, \asqa{} admits a clear notion of correctness that we use to define an overall metric of performance (\ovscore). Our empirical evaluations demonstrate that (i) \ovscore{} correlates well with the human judgment; and (ii) there is a large gap between human performance and the strong baselines. Thus, we believe that \asqa{} is an appealing task for the QA community.

Our analysis suggests that strong performance on \asqa{} is contingent upon both high-quality retrieval and summarization. These aspects constitute important directions for future work on \asqa{}. Additionally, an interesting direction for future work is to transfer our human annotation interface to a public crowdsourcing platform and ensure that crowdsourcing workers can reliably perform evaluations. This contribution would resolve the scalability problem of the expert-based approach to human evaluations we employ in this work.

%% file: sections/9appendix.tex
\medskip

\noindent We now provide additional discussion of several aspects of this work.

\section{Additional Details on the Annotation Procedure}
\label{appendix:annotation}

We begin with an additional discussion of the annotation procedure.

\paragraph{Construction of Context Paragraphs} As discussed in Section~\ref{section:preliminaries}, in our annotation task, we supplement each disambiguation $(\dq_i, \da_i)$ from \ambig{} with a context passage $C_i$. Let us now describe the procedure used to construct these context passages. 

For each disambiguation $(\dq_i, \da_i)$, we execute the following three-stage procedure:
\begin{enumerate}[topsep=5pt, itemsep=0pt, leftmargin=*]
    \item Among all paragraphs from Wikipedia pages $\wiki$ visited by \ambig{} annotators, we select those that contain $\da_i$.
    
    \item We compute TF-IDF similarity~\citep{sammut10tfidf} between the selected paragraphs and $\dq_i$.
    
    \item If the highest similarity exceeds a certain empirically selected threshold, we use the corresponding paragraph as an additional context $C_i$ provided to annotators. Otherwise, we do not provide context for that disambiguation ($C_i = \emptyset$).
\end{enumerate}

Following this procedure, we were able to provide non-empty additional context passages for 45\% of all disambiguations used in our annotation procedure. 

\paragraph{Quality Control and Feedback} Next, we discuss additional steps we took to help annotators in writing answers that satisfy the objectives formulated in Section~\ref{section:objectives}. First, we added an automated check to our interface that warns annotators if any of the short answers $\{\da_i\}_{i=1}^{\numdis}$ is missing from their long-form answer. Annotators were able to override the warning if they believe that an equivalent formulation of the missing short answer is already included. For example, given two disambiguations with short answers ``four seasons'' and ``4 seasons'', annotators were instructed to use any of these two equivalent options.

Second, in addition to the carefully designed training procedure, we were also continuously monitoring the annotators' performance as they were going through the task. In that, we were giving regular constructive feedback that highlighted areas of improvement and pointed out mistakes identified in annotators' past answers. While we did not observe any significant decay in quality between the exam session and the main task annotation, we believe that continuous monitoring is crucial to avoid creating an incentive for annotators to reduce the amount of effort they put into the task. 

Finally, to ensure that annotators did not have to guess when they met some situation not explained in the instructions, we maintained an FAQ document in which annotators could ask their questions and receive an answer within a day. To support this mechanism, we allowed annotators to ``park'' an annotation task they were unsure about and return to it after they have their concerns resolved.

\paragraph{Annotators' Well-Being} For this study, we recruited annotators who were fully dedicated to our task (8 hours a day for 5 days a week). To reduce the pressure on annotators and allow them to work at a comfortable pace, we gave annotators one hour to answer each question and recommended answering ten or more questions per day. On average, it took annotators 15 minutes to answer each question with the time consumption slightly decreasing as annotators get familiar with the task. The compensation rate for the task was set to be \$17.8/hour which is higher than the minimum hourly wage in the US.

\section{Additional Details on Modeling}
\label{appendix:model}

In this section, we provide additional details on the modeling aspect of our evaluations.

\begin{figure}[b]
    \centering
    \includegraphics[width=\linewidth]{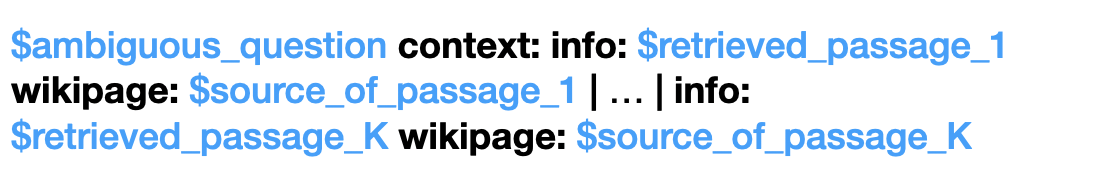}
    \caption{Input to the \tfob{K} model. }
    \label{fig:format_t5_0}
\end{figure}

\begin{figure}[h]
    \centering
    \includegraphics[width=\linewidth]{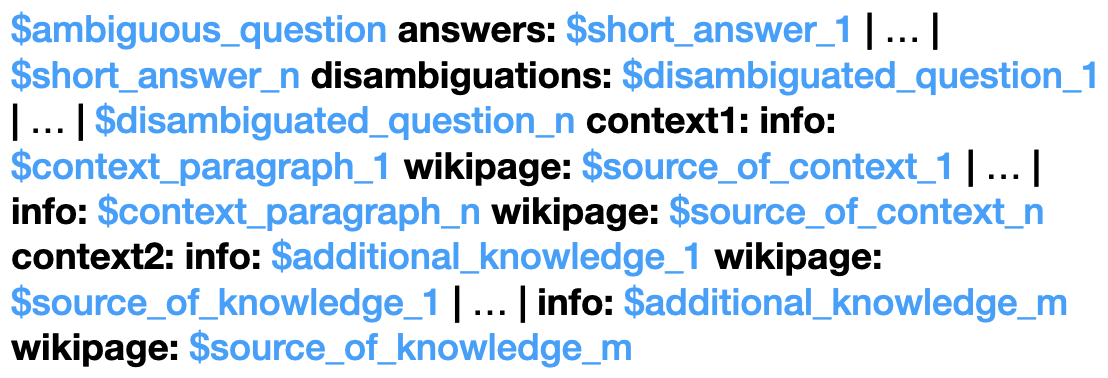}
    \caption{Input to the \oracle{} model.}
    \label{fig:format_t5_oracle}
\end{figure}

\paragraph{Input Format} Figures~\ref{fig:format_t5_0} and~\ref{fig:format_t5_oracle} provide schematic representations of inputs to the \tfob{K} and \oracle{} models, respectively. Bold black text represents tags that separate conceptually different parts of the input, text in blue is replaced with the instance-specific content in the actual training and evaluation data.

The input to \tfob{K} is simpler and consists of two parts separated by the \texttt{context} tag: an ambiguous question and $K$ retrieved passages. Each retrieved passage consists of the \texttt{info} field that contains the retrieved passage and the \texttt{wikipage} field that displays the title of the source Wikipedia page. Retrieved passages are separated with the pipe symbol ``|''.

The input to the \oracle{} model is more complex and has five parts:
\begin{itemize}[topsep=5pt, itemsep=0pt, leftmargin=*]
    \item An ambiguous question $q$
    \item Short answers $\{y_i\}_{i=1}^n$ (\texttt{answers})
    \item Disambiguated questions $\{x_i\}_{i=1}^n$\\ (\texttt{disambiguations})
    \item Context paragraphs $\{C_i\}_{i=1}^n$ (\texttt{context1})
    \item Additional knowledge pieces provided by the annotator $\{e_j\}_{j=1}^m$ (\texttt{context2})
\end{itemize}

Similarly to the \tfobna{} model, context paragraphs and additional knowledge pieces have \texttt{info} and \texttt{wikipage} fields, and the pipe symbol ``|'' is used to separate elements in the list.  

\paragraph{Parameter Choice} We use the context length of 512, 1024, and 2048 for the \tfob{1}, \tfob{3}, and \tfob{5} models, respectively. We use batch size of 8 across the three models. For \tfcb{}, we use a batch size of 16 with a context length of 128.